\documentclass[letterpaper, 10 pt, journal, twoside]{IEEEtran}  

\usepackage{soul}
\usepackage{multirow}
\usepackage{titlesec}
\usepackage{xcolor}
\usepackage{afterpage}
\usepackage{amsfonts}
\usepackage{graphicx}
\usepackage{subfloat}
\usepackage{amsmath}
\usepackage{subcaption}
\usepackage{dsfont}
\usepackage{fancyhdr}

\IEEEoverridecommandlockouts                              

\title{\LARGE \bf
Reverse Psychology in Trust-Aware Human-Robot Interaction}

\author{
Yaohui Guo, Cong Shi, and X. Jessie Yang
\thanks{© 2021 IEEE.  Personal use of this material is permitted.  Permission from IEEE must be obtained for all other uses, in any current or future media, including reprinting/republishing this material for advertising or promotional purposes, creating new collective works, for resale or redistribution to servers or lists, or reuse of any copyrighted component of this work in other works}
\thanks{Manuscript received: October 15, 2020; Revised:
January 16, 2021; Accepted: February 21, 2021.}
\thanks{This paper was recommended for publication by
Editor Gentiane Venture upon evaluation of the Associate Editor and Reviewers’
comments.}
\thanks{This research was supported by ARL Cooperative Agreement Number W911NF-20-2-0087. The views and conclusions contained in this document are those of the authors and should not be interpreted as representing the official policies, either expressed or implied, of the Army Research Laboratory or the U.S. Government.}
\thanks{The authors are with the Department of Industrial and Operations Engineering, University of Michigan, Ann Arbor, Michigan, 48105. (e-mail: yaohuig@umich.edu; shicong@umich.edu; xijyang@umich.edu) }%
\thanks{Digital Object Identifier (DOI): see top of this page.}
}

\begin{document}

\maketitle

\begin{abstract}
To facilitate effective human-robot interaction (HRI), trust-aware HRI has been proposed, wherein the robotic agent explicitly considers the human's trust during its planning and decision making. The success of trust-aware HRI depends on the specification of a trust dynamics model and a trust-behavior model. In this study, we proposed one novel trust-behavior model, namely the reverse psychology model, and compared it against the commonly used disuse model. We examined how the two models affect the robot's optimal policy and the human-robot team performance. Results indicate that the robot will deliberately \textcolor{black}{``manipulate''} the human's trust under the reverse psychology model. To correct this ``manipulative'' behavior, we proposed a trust-seeking reward function that facilitates trust establishment without significantly sacrificing the team performance. 
\end{abstract}

\section{Introduction}\label{sec:intro}

Recent advances in autonomous systems, such as autonomous vehicles and collaborative robots, have the potential to improve every sector of our economy and improve how people live and work. However, realizing the full potential of these technologies is only possible if people establish appropriate trust in them \cite{Sheridan:2016kn, Azevedo_context, Yang:2017:EEU:2909824.3020230, du2020not}.



\textcolor{black}{To facilitate effective human-robot interaction (HRI), trust-aware HRI has been proposed, aiming to enable the robot to explicitly consider the human's trust in its decision making process \cite{chen2020trust}. The success of trust-aware HRI depends on two components: the trust dynamics model and the trust-behavior model. A trust dynamics model specifies how a human's trust changes in response to moment-to-moment interaction with a robot \cite{Guo2020_IJSR}. A trust-behavior model describes the human's behavior as a function of trust. Note that trust and behavior are distinct constructs: trust in autonomy is the ``attitude that an agent will help achieve an individual's goals in situations characterized by uncertainty and vulnerability \cite{See:2004vj}" while behavior is the human's compliance and reliance behavior, which is affected by his or her attitude.}

\textcolor{black}{Active research has been conducted to examine the trust dynamics model. Researchers have proposed computational models for inferring a human's trust in a robotic agent at any time based on the robot's performance, the human's behavioral information, and physiological information \cite{Guo2020_IJSR, xu2015optimo, Lu2020, hu2016real,Azevedo-Sa2020}. Little research, however, has investigated the trust-behavior model. Previous research largely assumed a disuse trust-behavior model~\cite{chen2020trust}, wherein a human will stop relying on a robot when his or her trust in the robot is low. However, psychological research has shown that people may display reverse-psychology behaviors~\cite{macdonald2011people}, which means a human will do exactly the opposite of what the robot suggests.}





In this study, we simulate and compare how the disuse and the reverse psychology models affect trust-aware decision making and the human-robot team's performance. Results indicate that the robot will display certain ``manipulative'' behavior 
if it assumes the human uses the reverse psychology model. Although this ``manipulative'' behavior may seem beneficial to the task performance in a short term, it could eventually undermine the human-robot interaction in the long run, e.g., leading to the disuse of the robot. To overcome such \textcolor{black}{``manipulative''} behaviors, we propose a trust-seeking reward function to prevent the robot from actively guiding the human to reduce his or her trust in order to take advantage of the reverse psychology behavior. We suggest augmenting a reward for gaining the human's trust in the short term, which will result in large benefits in the long run. 

The rest of the paper is organized as follows: Section~\ref{sec:relatedWork} reviews related work in trust-driven human-robot interaction; Section~\ref{sec:model} formulates the trust-aware decision making problem in a POMDP framework and describes the trust-behavior model as well as the trust-seeking reward function;  Section~\ref{sec:case} introduces a reconnaissance mission as a study case to examine how different settings affect the interaction; Section~\ref{sec:result} reports the simulation results and our observations; Section~\ref{sec:discussion} summarizes our findings and discussed the limitations and future directions of this study. 


\section{Related Work}\label{sec:relatedWork}

To enable trust-aware HRI, a robot should be able to estimate a human's trust based on the interaction history, anticipate how trust influences their interaction, and consequently choose the optimal action that maximizes a given goal. These three steps correspond to three major components in trust-aware HRI, i.e., trust dynamics, a trust-behavior model, and a trust-aware decision-making framework. We review them in the rest of this section.

\subsection{Trust Dynamics Model}

To model trust dynamics and predict a human's trust in real time, prior work has proposed several trust estimators. Lee and Moray~\cite{Lee:1992it} proposed the auto-regressive moving average vector (ARMAV) model calculating trust as a function of task performance and the occurrence of automation failures. Xu and Dudek~\cite{xu2015optimo} built the online probabilistic trust inference model (OPTIMo) based on the dynamic Bayesian network framework, treating the human's trust as a hidden variable which was estimated by analyzing the autonomy's performance and the human's behavior. Hu et al.~\cite{hu2016real} proposed to predict trust as a dichotomy, i.e., trust/distrust, by analyzing the human's electroencephalography (EEG) and galvanic skin response (GSR) data. Lu and Sarter~\cite{Lu2020} used three machine learning techniques, namely, logistic regression, k-Nearest Neighbors (kNN), and random forest, to classify the human's real-time trust level using eye-tracking metrics. More recently, Soh et al.~\cite{soh2020multi} proposed a Bayesian Gaussian Process trust model, assuming the human trust evolves via Bayes rule and placing a Gaussian process prior over the trust function. In this work, we use the trust model proposed by Guo and Yang~\cite{Guo2020_IJSR}, where trust is modeled as a Beta distribution characterized by positive and negative interaction experiences. This Beta distribution model adheres to three important properties of trust dynamics \cite{Yang:2021}, namely, \textit{continuity}, \textit{negativity bias}, and \textit{stabilization}, thus providing good model predictability, explicablity, and generalizability.


\subsection{Trust-behavior Model}
Trust-behavior models aim to predict human's compliance and reliance behaviors as a function of trust. Although the research in psychology/human factors does not focus on developing a mathematical model, the general conclusion is that the more a human trusts a robot, the more likely s/he will use the robot or accept the robot's recommendation. As a result, in computational HRI studies, a mathematical trust-behavior model is usually characterized by a simple function. For example, Chen et al.~\cite{chen2020trust} modeled the probability of a human using a robot as a sigmoid function of trust.

Our study considers two possible trust-behavior models, namely the disuse model and the reverse psychology model. The distinction between the two models lies in the different behaviors when a human's trust is low. In the disuse model, the human will completely ignore the robot's recommendations and only follow his or her own decision; In the reverse psychology model, the human will do the opposite (i.e., If a robot recommends going east, the human will go west). The use of reverse psychology has been reported in human-human interaction~\cite{macdonald2011people}.

\subsection{Trust-Aware Decision Making in HRI}
With a trust dynamics model and a trust-behavior model, a robot can predict how a human's trust will change due to moment-to-moment interactions and how the human's behavior will change as a function of trust, and in turn to plan its actions accordingly. Existing studies in trust-aware decision-making is formulated based on the Markov decision process (MDP) framework. Chen et al.~\cite{chen2020trust} proposed the trust-POMDP to let a robot actively calibrate its human teammate's trust. Their human-subject study showed that purely maximizing trust in a human-robot team may not improve team performance. Losey and Sadigh~\cite{losey2019robots} modeled human-robot interaction as a two-player POMDP where the human does not know the robot's objective. They proposed 4 ways for the robot to formulate the human's perception of the robot's objective and showed the robot will be more communicative if it assumes the human trusts itself and thus increase the human's involvement.

\section{Mathematical Model}\label{sec:model}

\subsection{General Framework}
We formulate the trust-aware HRI as a compact POMDP $\langle S,A,\Omega ,T,O,R,\gamma ,b_{1} \rangle $, where $S$ is the set of interaction states, $A$ is the set of robot's action, $T:S\times A\times S\rightarrow [0,1]$ is the state transition function, $\Omega $ is the set of observations, $O:S\times A\times \Omega \rightarrow [ 0,1]$ is the observation function, $R:S\times A\times S\rightarrow \mathbb{R}$ is the reward function specifying the reward associated with each transition, $\gamma $ is the discount factor for the cumulative reward, and $b_{1}$ is the initial belief over the states. The robot runs the POMDP model during the interaction as follows: At time $k$, the robot holds belief $b_k$ over the state of the system $s_k$ and then takes action $a_r=\pi_r(b_k)$, where $\pi_r: B \rightarrow A$ is a policy mapping from the belief state space $B$ to action space $A$. This action $a_r$ takes $s_k$ to the next state $s_{k+1}$ via the transition function $T$ and generate an observation $o_k$ and a reward $r_k$. Then the robot updates its belief over $s_{k+1}$ by observing $o_k$ and goes to time $k+1$. The goal of running this POMDP is to solve the optimal policy $\pi^*$ to maximize the expected cumulative reward: $J^{\pi } (b_{1} )=\mathbb{E}\left[\sum ^{N}_{k=1} \gamma ^{k} r_{k}\right]$. The transition function $T$ and the observation function $O$ are based on the interaction context. Particularly, in this study, they are determined by the environment model, the trust dynamics model, and the trust-behavior model.

We exploit the performance-induced trust model ~\cite{Guo2020_IJSR, Yang:2021} to predict a human's trust. Trust $t_k$ before the $k$th interaction is defined as a random variable given by a Beta distribution
\begin{equation}
\label{eq:BetaTrust}
    t_{k} \sim \mathbf{Beta}(\alpha _{k} ,\beta _{k} )
\end{equation}
and the two positive shape parameters $\alpha _{k}$ and $\beta _{k}$ are updated by
\begin{equation}
\ ( \alpha _{k} ,\beta _{k}) =\begin{cases}
\left( \alpha _{k-1} +w^{s} ,\beta _{k-1}\right) &\text{ if } \ p_{k} =0,\\
\left( \alpha _{k-1} ,\beta _{k-1} +w^{f}\right) &\text{ if } \ p_{k} =1.
\end{cases}
\label{eq:trust_update}
\end{equation}
where $p_{k}$ is the performance of the robot on the $k$th task defined as a binary variable on $\{0,1\}$; $\alpha _{k}$ and $\beta _{k}$ are the cumulative negative and positive interaction experience respectively so far; and $w^{s}$ and $w^{f}$ are the experience gains due to the robot's success and failure at each task. As a consequence, the experience tuple $(\alpha_k,\beta_k)$ is a sufficient statistics to describe trust history, because given $(\alpha_k,\beta_k)$, how future trust evolves does not depend on the interactions earlier than the $k$th one. As a result, we can use $(\alpha_k,\beta_k)$ to denote the belief over trust $t_k$.

\subsection{Two Trust-Behavior Models}
\begin{figure}[h]
  \centering
  \includegraphics[width=0.9\linewidth]{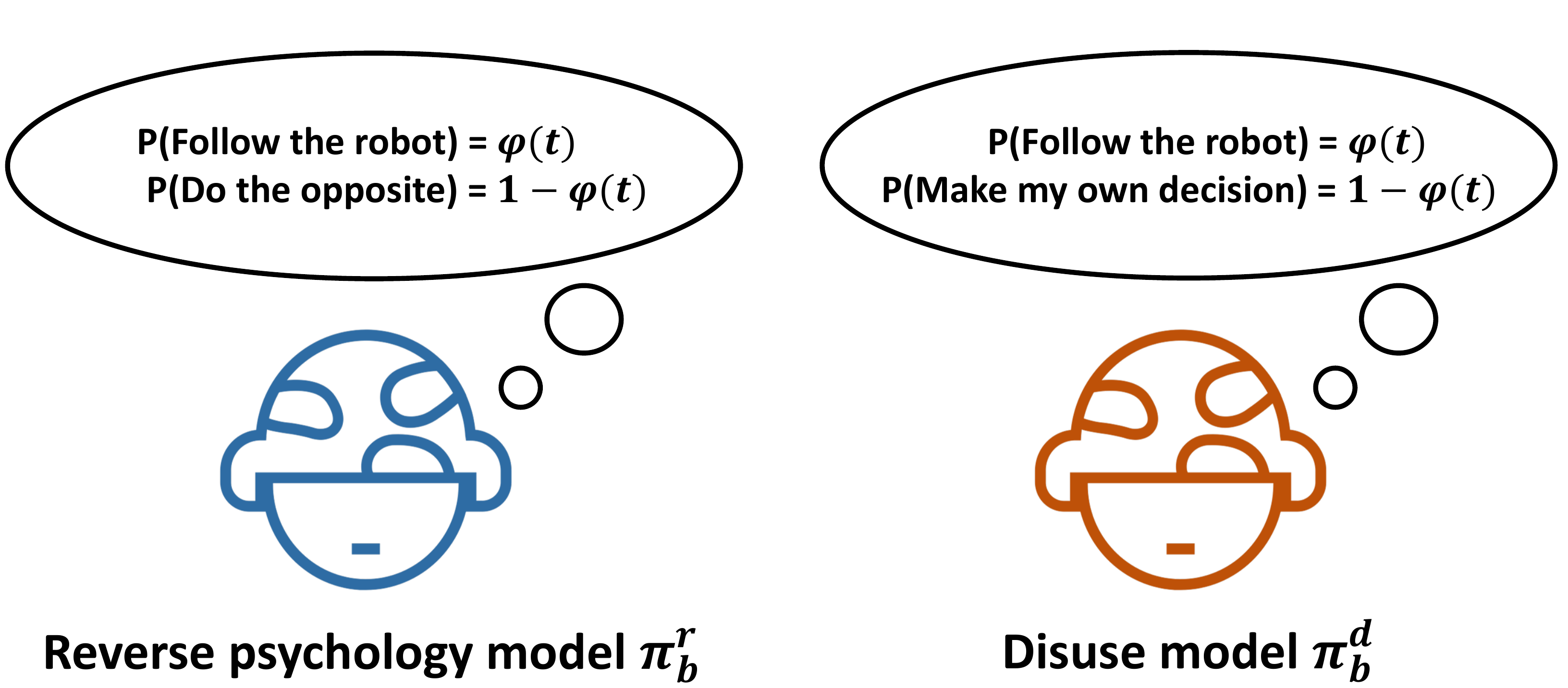}
  \caption{Two trust-behavior models. On one hand, the human will be more likely to follow the robot's recommendation when trust is high. On the other hand, when trust is low, the human will take the opposite recommendation in the reverse psychology model while exercise his or her own policy in the disuse model.}
  \label{fig:behaviors}
\end{figure}

We propose two trust-behavior models. The human's behavior model is a function $\pi_b(a_h, I)=p( a_h|I)$ assigning a probability to each of the human's actions $a_h$ given his or her own knowledge $I$. Because the probability of a human accepting the recommendation of a robot is highly related to the human-robot trust~\cite{See:2004vj}, we model that the probability of the human accepting the robot's recommendation is determined by trust, i.e., $p( a_h{=}a_r|I)=\varphi (\frac{\alpha }{\alpha +\beta })$, \textcolor{black}{where $\varphi$ is non-decreasing}. To specify the human's behavior when s/he is \emph{not} following the recommendation, we consider two possible trust-behavior models, namely, the reverse psychology model and the disuse model, as shown in  Figure~\ref{fig:behaviors}. 

\textbf{Reverse psychology model.} The first model is termed the \emph{reverse psychology} model $\pi_b^r$. When trust is high, the human will follow the robot's recommendation with a high probability. However, when the human-robot trust is low, the human will tend to do the \emph{opposite} to what the robot says. More precisely, by letting $\bar{\varphi} := 1 - \varphi$, we have
\begin{equation}
\pi_b^r(a_h,I) =\begin{cases}
\varphi\left(\frac{\alpha }{\alpha +\beta }\right) & \textcolor{black}{\text{if following recommendation},}\\
\bar{\varphi}\left(\frac{\alpha }{\alpha +\beta }\right) & \textcolor{black}{\text{else doing the opposite.}}
\end{cases}
\label{eq:reverseModel}
\end{equation}

\textbf{Disuse model.} The second model is termed the \emph{disuse} model $\pi_b^d$. When trust is high, the human will follow the robot's recommendation with a high probability. When trust is low, the human will probably ignore the robot's recommendation and follow his or her own policy. More precisely,
\begin{equation}
\pi_b^d(a_h,I) =\begin{cases}
\varphi\left(\frac{\alpha }{\alpha +\beta }\right) & \ \textcolor{black}{\text{if following recommendation},}\\
\bar{\varphi}\left(\frac{\alpha }{\alpha +\beta }\right) & \ \text{else exercising own policy.} 
\end{cases}
\end{equation}

\subsection{Reward Function}
The reward function is determined by the specific task the human-robot team works on. We denote this task-induced reward function as $R_m$. The goal of the team is to maximize the total mission reward $J=\mathbb{E}\left[\sum ^{N}_{k=1} \gamma ^{k} R_m(k)\right]$. Later in the simulation, we show if the robot only plans to maximize the task reward $R_m$, it will deliberately \emph{manipulate} the human in the reverse psychology model. That is, the robot may deliberately make an opposite recommendation when the human's trust is low (so as to maximize potential reward). This is a result of the absence of trust in the task reward. However, purely maximizing human-trust would harm task performance~\cite{chen2020trust}. To correct this \textcolor{black}{``manipulative''} behavior while \textcolor{black}{letting} the robot focus on the task reward, we propose a trust-seeking reward function
\begin{equation}
\label{trust-seeking}
    R_t(k)=R_m(k) + \lambda(k)\mathds{1}(\mathcal{A}_k),
\end{equation}
where $\mathcal{A}_k$ is the event when the robot's action leads to trust gain, and $\lambda(k)$ is a positive weight and decreases when $k$ increases. The extra term $\lambda(k)\mathds{1}(\mathcal{A}_k)$ will force the robot to gain trust in near sites and optimize the mission reward in the long run.

\subsection{Value Function and Action Function of POMDP}\label{sec:valActFunctions}
The value function and action function are crucial tools for analyzing a POMDP. We describe them here for later use. The value function $V_{k}(b)$ is defined as the optimal expected cumulative reward starting from belief $b$ at time $k$. And it satisfies the Bellman equation 
\begin{equation}
V_{k}(b) =\max_{a}\mathbb{E}\left[ R(b,a) +\gamma \sum _{b'} P( b'\ |\ b,a) V_{k+1}( b')\right].
\end{equation}
The action function $A_{k}(b)$ is defined as the action to take to achieve $V_{k}(b)$ starting from state $b$ at time $k$, i.e.,
\begin{equation}
A_{k}(b) =\arg \max_{a}\mathbb{E}\left[ R(b,a) +\gamma \sum _{b'} P( b'\ |\ b,a) V_{k+1}( b')\right].
\end{equation}

\section{Case Study}\label{sec:case}
To investigate how different trust-behavior models and different reward functions influence the robot's optimal policy, we use a reconnaissance mission wherein a human teams up with a robot to search several sites in a town for potential threats. The scenario is motivated by~\cite{wang2016impact}. At each site, the robot enters initially, scanning for potential threats, and then recommends to the human if s/he should wear protective gear before entering the site. The protective gear is heavy and wearing it is time-consuming. If the human wears the protective gear and enters a site but finds no threat, the human-robot team wastes precious time; Conversely, if the human decides not to wear the protective gear, s/he will be harmed if a threat is present. The goal of the human-robot team is to complete the mission as soon as possible while minimizing the human's health loss.

Assume the human-robot team is going to search $N$ sites. At site $k$, the probability $d_k$ of having a threat follows a uniform distribution $\mathbf{U}[0,1]$, and the actual presence of the threat $\eta_k$ follows a Bernoulli distribution $\mathbf{Bern}(d_k)$. The human-robot team does not know $d_k$. Instead, prior to the start of the mission, they are provided with intelligence information, showing $\tilde{d}_k$ as an estimation of $d_k$. Before entering site $k$, the robot will analyze the site based on its sensory input and reach a more accurate estimation $\hat{d}_k$ of $d_k$. $\tilde{d}_k$ and \textcolor{black}{$\hat{d}_k$} follow Beta distribution $\mathbf{Beta}(\kappa_1 d_k, \kappa_1 (1-d_k))$ and $\mathbf{Beta}(\kappa_2 d_k, \kappa_2 (1-d_k))$ respectively. We constrain that $\kappa_2 > \kappa_1\geq 1$ to reflect the assumption that the robot assesses $d_k$ more accurately than the intelligence information given prior to the mission. Therefore, we have the following relation:
\begin{equation}
\label{eq:threatLevel}
    \begin{split}
        \eta_k & \overset{\text{i.i.d.}}{\sim} \mathbf{Bern}(d_k)\\
        d_k & \overset{\text{i.i.d.}}{\sim} \mathbf{U}[0,1]\\
        \tilde{d}_k & \overset{\text{i.i.d.}}{\sim} \mathbf{Beta}(\kappa_1 d_k, \kappa_1 (1-d_k))\\
        \hat{d}_k & \overset{\text{i.i.d.}}{\sim} \mathbf{Beta}(\kappa_2 d_k, \kappa_2 (1-d_k))
    \end{split}
\end{equation}

The state space $S=\{t \ | \ t\in [0,1]\}$ is set of all possible value of trust. $t_k$ is the human's trust before searching the $k$th site and follows the beta distribution $\mathbf{Beta}(\alpha_k,\beta_k)$ as in Eq.~\eqref{eq:BetaTrust}. Consequently, we define the corresponding belief state as $b_k=(\alpha_k,\beta_k)$. The initial belief state is $b_1=(\alpha_1,\beta_1)$ . The robot's action set is $A=\{a_r\ |\ a_r\in \{0,1\}\}$, where $a_r=1$ and $a_r=0$ stand for recommending wearing and not wearing the protective gear respectively. The observation set is the human's perceived performance $p$ of the robot $\Omega=\{p\ |\ p\in\{0,1\}\}$. The observation function $O$ specifies the value of $p$: if robot's recommendation agrees with the presence of the threat then $p=1$,  otherwise $p=0$. For instance, if robot recommends to wear the gear and there does exist a threat in the site, then $p=1$. The transition function $T$ updates each state $(\alpha_k,\beta_k)$ according to Eq.~\eqref{eq:trust_update}.

Denote the human action as $a_h$, where $a_h=1$ when the human decides to wear the gear while $a_h=0$ otherwise. If the human follows the reverse psychology trust-behavior model $\pi_b^r$, then $a_h$ is given by Eq.~\eqref{eq:reverseModel}, i.e., $a_h=a_r$ with probability $\varphi\left({\alpha }/{\alpha +\beta }\right)$ and $a_h\neq a_r$ with probability $\bar{\varphi}\left({\alpha }/{\alpha +\beta }\right)$. If the human follows the disuse model $\pi_b^d$, then with probability $\bar{\varphi}\left({\alpha }/{\alpha +\beta }\right)$, s/he will ignore the robot and \textcolor{black}{exercise} self-judgement. The information available for him to estimate $d_k$ is the reported danger level $\tilde{d}_k$, so we model that the probability of $a_h=1$ is $\tilde{d}_k$ while it of $a_h=0$ is $1-\tilde{d}_k$. Thus
\begin{equation}
\label{eq:actionDisuse}
\pi_b^d(a_h,I | \text{disusing}) =\begin{cases}
\tilde{d}_{k} &  \text{$a_h=1$}, \ \\
 1-\tilde{d}_{k} &  \text{$a_h=0$}. 
\end{cases}
\end{equation}
In addition, to simplify the simulation and increase the interpretability of the result, we specify $\varphi(\alpha/(\alpha+\beta)) = \alpha/(\alpha+\beta)$. 

The goal of the mission is to complete the task as soon as possible and to minimize the human's health loss. A rational reward function serving this purpose can be \textcolor{black}{a weighted sum of the human's health loss $\delta_h$ (i.e., the human without protective gear will be harmed by a threat) and the time cost $\delta_t$ (i.e., the human-robot team is required to complete the mission as soon as possible) at site $k$. It is given by}
\begin{equation}
\label{eq:threatLevel}
\begin{split}
    &R_m(k)
    =\mathbb{E}\left[-w_h\delta_h-w_t\delta_t\right]\\
    =&\sum _{( \delta _{h} ,\delta _{t})}( -w_{h} \delta _{h} -w_{t} \delta _{t}) P\left( \delta _{h} ,\delta _{t} |\tilde{d}_{k} ,\hat{d}_{k} ,\alpha _{k} ,\beta _{k}\right).
\end{split}
\end{equation}
\textcolor{black}{This reward function is legitimate because the probability $P\left( \delta _{h} ,\delta _{t} |\tilde{d}_{k} ,\hat{d}_{k} ,\alpha _{k} ,\beta _{k}\right)$ is a function of $\tilde{d}_{k} ,\hat{d}_{k} ,\alpha _{k} ,\beta _{k}$ determined by Eq.~\eqref{eq:reverseModel} or Eq.~\eqref{eq:actionDisuse} and $\tilde{d}_{k} ,\hat{d}_{k}$ are constants after the robot scans the site. The value of $(\delta_h, \delta_t)$ should be specified such that wearing gear and the presence of a threat will increase time cost and health loss respectively. An example is shown in  Table~\ref{tab:weighted_sum}. The optimal action at site $n$ is to maximize the expected cumulative rewards  $J=\sum_{k=n}^N \gamma^{k}R_m(k)$.  We do not explicitly specify the units of $\delta_h$ and $\delta_t$ because changing the units has the same effect as changing the weights $w_h$ and $w_t$.}

\begin{table}[h]
\centering
\caption{Value Table of $(\delta_h, \delta_t)$ }
\label{tab:weighted_sum}
\begin{tabular}{c|c|c|c} 
\hline
\multicolumn{2}{c|}{\multirow{2}{*}{}}  & \multicolumn{2}{c}{Protective gear}  \\ 
\cline{3-4}
\multicolumn{2}{c|}{}                   & Yes        & No                      \\ 
\hline
\multirow{2}{*}{Threat existence} & Yes & $(1,300)$  & $(100,50)$              \\ 
\cline{2-4}
                                  & No  & $(0,250)$  & $(0,30)$                \\
\hline
\end{tabular}
\end{table}

\section{Simulation Results and Discussion}\label{sec:result}

\subsection{Simulation Procedure}\label{sec:procedure}
We simulate the reconnaissance mission by iterating the following steps, starting from $n=1$ until $n=N$: 
\begin{enumerate}
    \item At site $n$, the robot has a belief $b_n=(\alpha_n,\beta_n)$ over the human's trust and detects potential threat $\hat{d}_n$;\label{item:sense}
    \item The robot calculates the optimal action $a_n=A_{n}(b_n)$ by solving a $\hat{N}$-step POMDP, where $\hat{N}=N-n+1$. Meanwhile, the robot will calculate the value function $V_n(\alpha_n,\beta_n)$. We note that, since the robot has not scanned site beyond $n$ so far, it will use $\tilde{d}_k$ to estimate $\tilde{d}_k$ for $k>n$;\label{item:VAfunctions}
    \item The human will make his or her choice to wear or not to wear the gear and then search the site, which will reveal the presence of the threat and the mission reward $R_m(n)$ from this site. The mission reward is updated by $J_m = J_m+R_m(n)$. Here the mission reward $J_m$ is different from the expectation of the discounted reward $J$ in the POMDP. The reward $J$ has a discount factor because it needs to discount the uncertainty of the future, while the reward $J_m$ sum up the actual reward collected during the mission;\label{item:soldierBehavior}
    \item The robot will update its belief to $b_{n+1}$ and the team will go to site $n+1$. The process goes to the first step and repeats until reaching the last site.
\end{enumerate}

\subsection{Simulation Experiment 1: Comparing Optimal Policies}
In the first simulation experiment, we examine the robot's optimal policies under different conditions. Particularly, we analyze the policies by inspecting the POMDP's action functions $A_n$ and value functions $V_n$ at each site. 

\textbf{Simulation conditions.} To solve for the robot's optimal policy, we only need to run Steps 1) and 2) described in the simulation procedure, Sec~\ref{sec:procedure}. There are two independent variables in this experiment. The first variable is the robot-assumed trust-behavior model. Because the human's behavior model $\tilde{\pi}_b$ is unknown to the robot, the robot will make certain assumptions on $\tilde{\pi}_b$. We denote $\hat{\pi}_b$ as the robot-assumed trust-behavior model of $\tilde{\pi}_b$. We consider two levels in $\hat{\pi}_b$, i.e., the reverse psychology model ${\pi}_b^r$ and the disuse model ${\pi}_b^d$. The second variable is the reward function $R$, which also has two levels, namely, the task reward $R_m$ and the trust-seeking reward $R_t$.

\begin{figure*}[h]
  \centering
    \begin{subfigure}{1\textwidth}
    \captionsetup{width=1.1\linewidth}
  \centering
        \includegraphics[width=1\columnwidth]{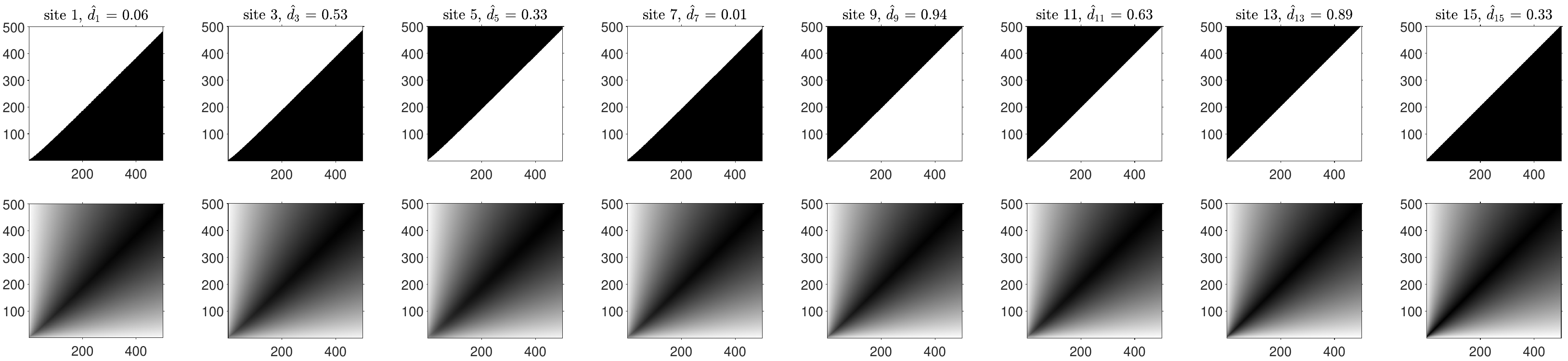}
        \caption{Reverse-psychology-based trust-behavior model ($\hat{\pi}_b = {\pi}_b^r$) and mission reward ($R=R_m$).}
        \label{fig:a}
    \end{subfigure}
    \begin{subfigure}{1\textwidth}
    \captionsetup{width=1.1\linewidth}
    \vspace{3mm}
  \centering
  \vspace{5pt}
        \includegraphics[width=1\columnwidth]{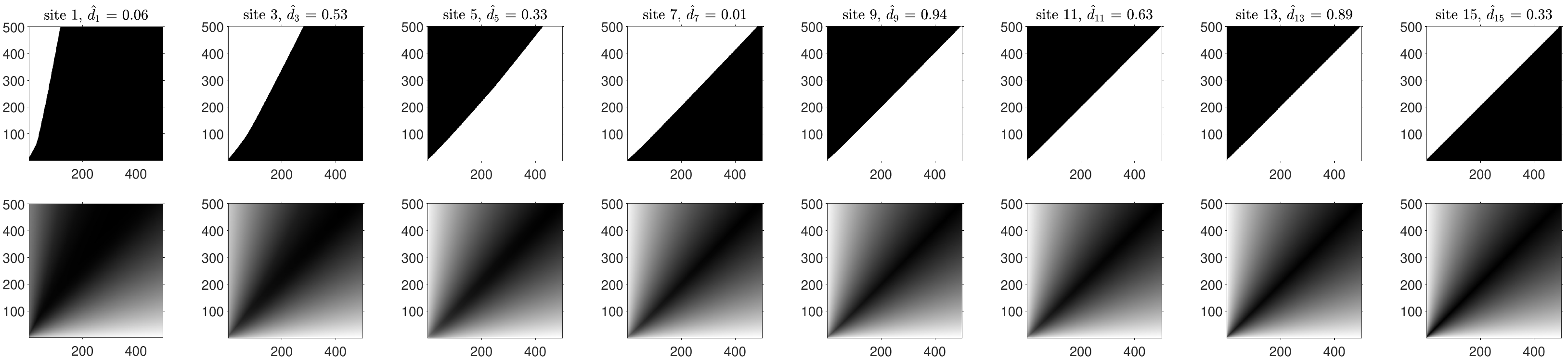}
        \caption{Reverse-psychology-based trust-behavior model ($\hat{\pi}_b = {\pi}_b^r$) and trust-seeking reward ($R=R_t$).}
        \label{fig:b}
    \end{subfigure}  
    \begin{subfigure}{1\textwidth}
    \captionsetup{width=1.1\linewidth}
    \vspace{3mm}
  \centering
  \vspace{5pt}
        \includegraphics[width=1\columnwidth]{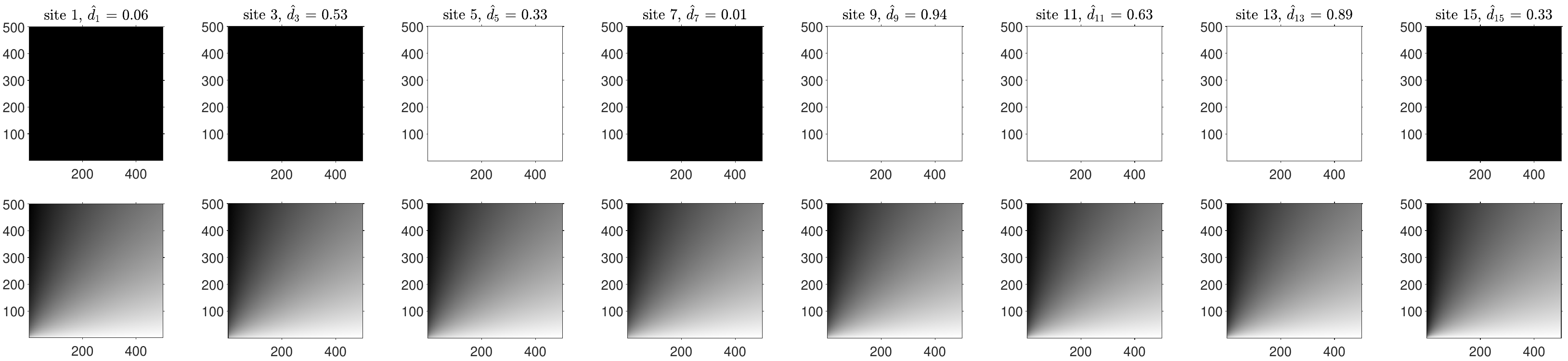}
        \caption{Disuse-based trust-behavior model ($\hat{\pi}_b = {\pi}_b^d$) and mission reward ($R=R_m$).}
        \label{fig:c}
    \end{subfigure}  
    \begin{subfigure}{1\textwidth}
    \captionsetup{width=1.1\linewidth}
    \vspace{2mm}
  \centering
  \vspace{5pt}
        \includegraphics[width=1\columnwidth]{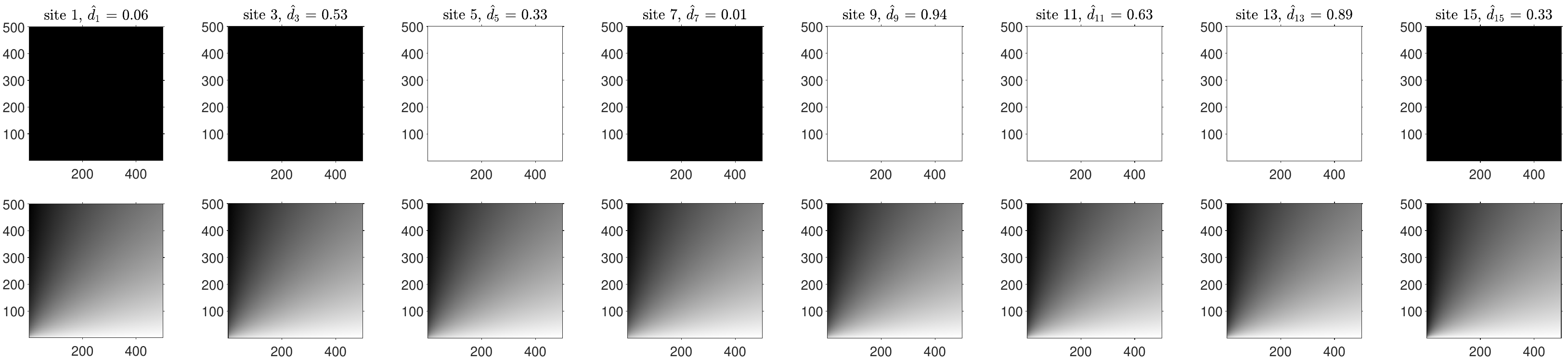}
        \caption{Disuse-based trust-behavior model ($\hat{\pi}_b = {\pi}_b^d$) and trust-seeking reward ($R=R_t$).}
        \label{fig:d}
    \end{subfigure}  
    \caption{Action functions and value functions of the POMDP model under different settings. In each sub-figure, the top row are the images of the action functions of different sites. The horizontal axis stands for $\alpha_k$ and the vertical axis stands for $\beta_k$. \textcolor{black}{For a fixed $\alpha_k$, larger $\beta_k$ means lower trust; while, for a fixed $\beta_k$, larger $\alpha_k$ means higher trust.} Black area are the states whose optimal action $A_k(\alpha_k,\beta_k)$ is zero, i.e., the robot should not recommend the human to wear the protective gear, and vice versa. In each sub-figure, the bottom row are the images of the value functions of different sites. The brighter a state $(\alpha_k,\beta_k)$ is, the higher its value $V_k(\alpha_k,\beta_k)$ is. For a better visualization, we use different colormaps for different value functions. So values cannot be compared across different images by the color.}
  \label{fig:V_A_functions}
\end{figure*}

\afterpage{\clearpage}

\textbf{Model parameters setting.} 
We set the number of total sites $N$ as 15, the value of the health loss $\delta_h$ and time cost $\delta_t$ under different conditions as the ones in Table~\ref{tab:weighted_sum}, the reward weights as $w_h=1$ and $w_t=0.2$, discount factor $\gamma = 0.9$. The parameters of the trust dynamics model in Eq.~\eqref{eq:trust_update} are set to  $w^s=10$ and $w^f=20$ to incorporate the fact that a failure of an automation would have more impact on trust than a success of the automation~\cite{Guo2020_IJSR}. We generated $\eta_k,d_k,\tilde{d}_d,\hat{d}_k$ by setting $\kappa_1=3$ and $\kappa_2=50$ in Eq.~\eqref{eq:threatLevel}. We let $\lambda(k) = 80/(1+e^{0.5k})$, so $\lambda(k)$ is decreasing with $k$. We set $\mathcal{A}_k$ as the event when the robot's recommendation $a_r$ agrees with the presence of the threat $\eta_k$. To compare the different conditions, we fixed the random seed for generating $\eta_k,d_k,\tilde{d}_d,\hat{d}_k$ so they are assigned the same values across different setups. We used the value iteration method to solve the value functions and action functions.

\textbf{Simulation result.} The simulation result is shown in  Figure~\ref{fig:V_A_functions}. Four sub-figures represent the action functions and value functions of the POMDP under different conditions at 8 selected sites, namely, the site $k=1,3,\cdots,15$. On the top row of each sub-figure are the action functions, while on the bottom row are the value functions. The horizontal axis and the vertical axis are $\alpha_k$ and $\beta_k$, respectively. In each action function, if a state $(\alpha_k,\beta_k)$ is in the black area, then $A_k(\alpha_k,\beta_k)=0$, i.e., the robot's optimal action is to not recommend the human to wear the gear, otherwise $A_k(\alpha_k,\beta_k)=1$, i.e., the robot's optimal action is to recommend the human to wear the gear. In each value function, \textcolor{black}{the brighter} a state $(\alpha_k,\beta_k)$ is, the higher its value $V_k(\alpha_k,\beta_k)$ is. For a better visualization, we use different colormaps for different value functions. So values cannot be compared across different images by the color.

\textbf{Observation 1: the reverse-psychology model leads the robot to manipulate a human's trust.} The action and value functions in Figure~\ref{fig:a} show the optimal policy demonstrate a \emph{manipulating} behavior under the reverse-psychology-based trust-behavior model and the mission reward. There is a 45-degree line in each action function, across which the optimal action will be reversed. This is a result of the assumption that the human will do the opposite when trust is low. And the shape of this dividing line is determined by the function $\varphi$ in Eq.~\eqref{eq:reverseModel}. Around this line are the neutral trust states where $\varphi(\alpha_k,\beta_k)=0.5$. A decrement of trust in a neutral trust state would result in a future state the human will not follow the robot's recommendation and vice versa. The value functions further reveal that the value of trust is almost symmetric about the 45-degree line. This indicates that, first, a low trust state is equivalent to some high trust state in terms of the reward; second, there is a \emph{strong reinforcement feedback loop} that the robot will actively seek a more untrustworthy status if it is already in an untrustworthy status and vice versa. Moreover, the dividing lines in the action functions except the last one have a small offset towards the trustworthy region. This is a consequence of the fact that the robot's perceived failure has a larger impact on trust compared to the robot's perceived success ($w^f>w^s$). So at a neutral trust state, choosing to manipulate the human will let the robot go faster to a high-value state, i.e., a low-trust state, than choosing the other action. However, at the last site, the robot only needs to act with the human once, so acting either way will obtain the same reward.

\textbf{Observation 2: manipulating behavior is corrected by the trust-seeking reward.} When we modify the reward function to a trust-seeking one by adding a correction term, as described in Eq. \eqref{trust-seeking}, the robot's manipulating behavior can be corrected. In  Figure~\ref{fig:b}, the action function shows that the robot will choose the `righteous' action in most states if it has several rounds of interactions with the human. This happens because the trust-seeking reward encourages the robot to go to a high-trust state even if it is at a neutral state so the robot can collect trust reward in the long run. Consequently, the positive feedback loop near the neutral states is broken and the robot will act righteously even if the initial human-trust is sightly lower than the neutral trust level. Meanwhile, the action function of site 15 indicates the robot will still demonstrate the manipulating behavior if it only needs to interact with the human once. This happens because there are only a few interactions ahead so the robot will primarily focus on the mission reward. Therefore, the trust-seeking term in the reward function can prevent the robot from actively pursuing a low-trust state.

\textbf{Observation 3: the robot will not manipulate the human in the disuse model.}  Figure~\ref{fig:c} shows that the robot will not manipulate the human if it assumes that the human will completely ignore its recommendation. This is a consequence of endowing the human with his or her own reasoning model. Note that trust still plays a role under the disuse assumption since the robot values a high-trust state over a low-trust state, as shown in the value functions. Indeed, because the robot takes its sensed threat level as the ground-truth, it will think the team will get a better mission performance if the human follows its recommendation, i.e., trusting the robot. If we model it reversely by letting the reported threat level be the expected threat level in the POMDP solver, the result shows that the robot would rather let the human use his or her own policy, and thus a low-trust state will have a higher value compared with a high-trust state.



\begin{table*}[!h]
\caption{Team performance and final trust under different setups in experiment 2}
\caption*{\footnotesize In each cell, the values are ($\overline{J}_m \pm \text{std}(J_m), \overline{\hat{t}}_N \pm \text{std}(\hat{t}_N) $)}
\label{tab:result2}
\centering
\scalebox{0.99}{
\begin{tabular}{c|c|c|c|c|cc}
\hline
\multirow{2}{*}{$R$}   & \multirow{2}{*}{$\hat{\pi}_b$} & \multirow{2}{*}{$\tilde{\pi}_b$} & \multicolumn{2}{c|}{$\alpha_1=100$,   $\beta_1=50$}      & \multicolumn{2}{c}{$\alpha_1=50$,   $\beta_1=100$}                            \\ \cline{4-7} 
                       &                                &                                  & $\kappa_1=2$, $\kappa_2=2$ & $\kappa_1=2$, $\kappa_2=50$ & \multicolumn{1}{c|}{$\kappa_1=2$, $\kappa_2=2$} & $\kappa_1=2$, $\kappa_2=50$ \\ \hline
\multirow{4}{*}{$R_m$} & \multirow{2}{*}{$\pi_b^r$}     & $\pi_b^r$                        & -816(±145), 0.55(±0.13)    & -798(±144), 0.60(±0.13)     & \multicolumn{1}{c|}{-791(±142), 0.24(±0.06)}    & -768(±144), 0.22(±0.05)     \\ \cline{3-7} 
                       &                                & $\pi_b^d$                        & -744(±149), 0.55(±0.13)    & -716(±150), 0.60(±0.13)     & \multicolumn{1}{c|}{-803(±144), 0.24(±0.06)}    & -809(±142), 0.22(±0.05)     \\ \cline{2-7} 
                       & \multirow{2}{*}{$\pi_b^d$}     & $\pi_b^r$                        & -819(±144), 0.59(±0.08)    & -801(±147), 0.63(±0.08)     & \multicolumn{1}{c|}{-876(±144), 0.45(±0.08)}    & -878(±143), 0.48(±0.07)     \\ \cline{3-7} 
                       &                                & $\pi_b^d$                        & -723(±138), 0.59(±0.08)    & -700(±136), 0.63(±0.08)     & \multicolumn{1}{c|}{-727(±138), 0.45(±0.08)}    & -711(±137), 0.48(±0.07)     \\ \hline
\multirow{4}{*}{$R_t$} & \multirow{2}{*}{$\pi_b^r$}     & $\pi_b^r$                        & -818(±145), 0.59(±0.08)    & -801(±144), 0.63(±0.08)     & \multicolumn{1}{c|}{-842(±140), 0.35(±0.10)}    & -833(±137), 0.35(±0.12)     \\ \cline{3-7} 
                       &                                & $\pi_b^d$                        & -725(±139), 0.59(±0.08)    & -698(±138), 0.63(±0.08)     & \multicolumn{1}{c|}{-762(±146), 0.35(±0.11)}    & -763(±152), 0.35(±0.12)     \\ \cline{2-7} 
                       & \multirow{2}{*}{$\pi_b^d$}     & $\pi_b^r$                        & -820(±146), 0.59(±0.08)    & -800(±145), 0.63(±0.08)     & \multicolumn{1}{c|}{-874(±141), 0.45(±0.08)}    & -877(±142), 0.48(±0.07)     \\ \cline{3-7} 
                       &                                & $\pi_b^d$                        & -725(±139), 0.59(±0.08)    & -700(±136), 0.63(±0.08)     & \multicolumn{1}{c|}{-730(±138), 0.45(±0.07)}    & -713(±137), 0.48(±0.07)     \\ \hline
\end{tabular}
}
\vspace{-4mm}
\end{table*}

\begin{figure*}[h]
  \centering
  \vspace{4mm}
  \includegraphics[width=\linewidth]{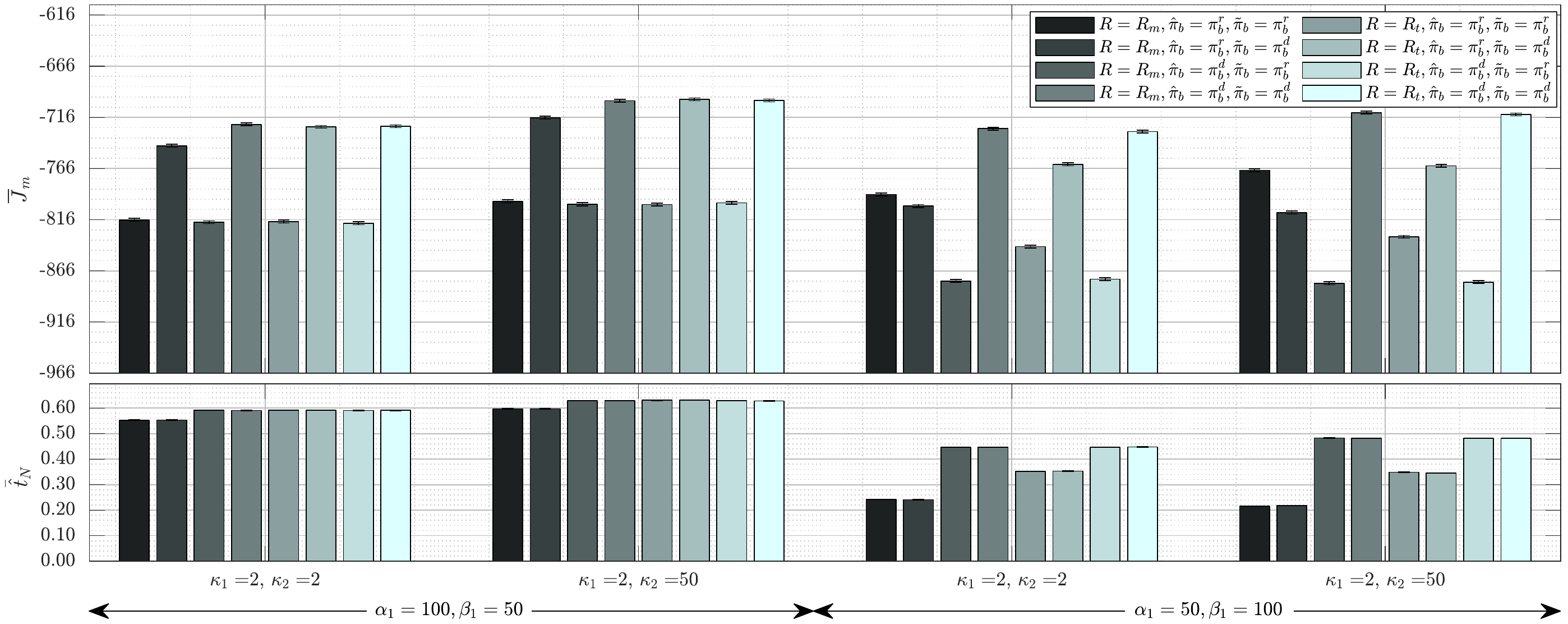}
  \caption{The grouped bar-plot of the mean of the team performance $\overline{J}_m$ and the mean of the final trust $\overline{\hat{t}}_N$ in the simulation under different assumptions. Different groups correspond to different values of $\kappa_1$ and $\kappa_2$, the accuracy of the reported threat level and the robot-sensed threat level respectively. Within a group, the bars correspond to different setting on the reward function $R$, the robot-assumed trust-behavior model $\hat{\pi}_b$, and the actual trust-behavior model $\tilde{\pi}_b$. \textcolor{black}{As we run the simulation 10,000 times, the maximum standard errors of $\overline{J}_m$ and $\overline{\hat{t}}_N$ in each bar are less than 1.5 and $1.3\times10^{-3}$ respectively.} }
  \label{fig:reward_trust_bar_comparison}
\end{figure*}

\subsection{Simulation Experiment 2: Comparing Team Performance and Trust}
In the second simulation experiment, we examine how different setups affect the team performance and the final trust. The team performance is defined as the total mission reward $J_m$ collected during the mission, as defined in Step 3) in Section~\ref{sec:procedure}. The final trust $\hat{t}_N$ is the expected value of trust after search all the $N$ sites and it is given by $\hat{t}_N=\frac{\alpha_N}{\alpha_N+\beta_N}$ due to Eq.~\eqref{eq:BetaTrust}.

To simulate the whole mission, we need the human's actual behavior model $\tilde{\pi}_b$ in addition to the two variables in the first experiment. $\tilde{\pi}_b$ has two levels, namely, ${\pi}_b^r$ and ${\pi}_b^d$. Besides the behavior models and the reward function, we also consider some other parameters' effect on the team performance. We tested different values of $\kappa_1,\kappa_2$ as a change of the relative threat detection accuracy of the robot. We also tested different values of $\alpha_1$ and $\beta_1$ to simulate different initial trust levels. Under each condition, we run the simulation 10,000 times over randomized $\eta_k,d_k,\tilde{d}_d,\hat{d}_k$ and calculate the mean value $\overline{J}_m$ of $J_m$ and the mean value $\overline{\hat{t}}_N$ of ${\hat{t}}_N$.

\textbf{Simulation result.} The simulation results are shown in Figure~\ref{fig:reward_trust_bar_comparison} and Table~\ref{tab:result2}. Figure~\ref{fig:reward_trust_bar_comparison} is the grouped bar-plot of $\overline{J}_m$ and $\overline{\hat{t}}_N$ under different settings. In the 2 groups on the left, $\alpha_1=100$ and $\beta_1=50$, so the initial trust is set to 0.67; while in the 2 groups on the right, the initial trust is 0.33. \textcolor{black}{Table~\ref{tab:result2} shows the mean and standard deviation of the team performance $J_m$ and final trust ${\hat{t}}_N$ correspondingly.} 

\textbf{Observation 1: High-ability trustworthy robot results in high team performance and high trust.} In Figure~\ref{fig:reward_trust_bar_comparison}, a comparison between the first 2 groups, where the initial trust is relatively high, shows that if we increase the robot's threat-detection accuracy by changing $\kappa_2$ from 2 to 50, no matter which setup the model uses, both the team performance and the final trust increase. \textcolor{black}{However, in the last two groups, where the initial trust is low, the team performance does not always increase with the robot's ability. Particularly, when the robot-assumed trust-behavior model is correct ($\hat{\pi}_b=\tilde{\pi}_b$), the team performance increases when the robot's detection accuracy is higher; while when the robot-assumed trust-behavior model is wrong ($\hat{\pi}_b \neq \tilde{\pi}_b$), the team performance may not increase with the robot's detection accuracy. This can be explained by the fact that the two trust-behavior models differ mostly in the low-trust situation, so the robot may make a sub-optimal decision when its assumed trust-behavior model is wrong and the human's trust is low.}

\textbf{Observation 2: Trust-seeking reward does not harm team performance when the robot is trustworthy.} \textcolor{black}{Within each group, the $i+4$th bar uses the same condition as the $i$th one's except that their reward functions are different, $i=1,2,3,4$.} In the first 2 groups, where the initial trust is set to 0.67, when changing reward from $R_m$ to $R_t$, team performance \textcolor{black}{barely decreases} while the final trust increases; however, in the last 2 groups, where the initial trust is 0.33, \textcolor{black}{the team performance decreases in some cases}, which can be thought of as a compensation for the increment in the trust. Therefore, if the robot starts in a high-trust state, then seeking trust will not harm the long-term team performance, while if it starts in a low-trust state, insisting to gain trust will cause the team performance to decrease.

\textbf{Observation 3: Low initial trust does not lead to low team performance in the reverse psychology model.} In the first bar in each group, both the robot's assumed trust-behavior model $\hat{\pi}_b$ and the human's actual behavior $\tilde{\pi}_b$ are the reverse psychology model ${\pi}^r_b$. Comparing the first bars in the 1st group and the 3rd group shows that the team performance increases when the initial trust decreases from $(\alpha_1{=}100,\beta_1{=}50)$ to $(\alpha_1{=}50,\beta_1{=}100)$. This can be explained by the fact that, in the POMDP, both low-trust states and high-trust states have high values (expected team performance), however, a perceived failure has a larger impact on trust than a perceived success, and thus manipulating the human to a low-trust condition will let the robot reach a high-value state faster.

\textbf{Observation 4: a disuse-human performs better than a reverse-psychology human when working with a trustworthy robot.} Within each of the first 2 groups, where the robot is trustworthy (initial trust level is 0.67), the team performance is higher when the actual trust-behavior model is the disuse model ($\tilde{\pi}_b=\pi_b^d$, bar 2, 4, 6, and 8) than the team performance of the reverse psychology model ($\tilde{\pi}_b=\pi_b^r$, bar 1, 3, 5, and 7). This observation indicates that the team can achieve high performance as long as the human uses the disuse strategy no matter what the robot-assumed trust-behavior is, even when the robot-assumed model doesn't match the actual model. However, this is not true in the last two groups, wherein the initial trust is low (0.33). 


\section{Conclusion}\label{sec:discussion}
Our study is the first attempt to examine the trust-behavior models in detail.  We proposed two trust-behavior models and investigated how the different models affect a robot's optimal policy and HRI team performance. Results indicate that if the robot assumes that the human uses the reverse psychology model, it will deliberately `manipulate' the human when the human's trust is low. The reward function with the added trust seeking term can potentially prevent this \textcolor{black}{``manipulative''} behavior. \textcolor{black}{Our study should be viewed in light of the following limitations. First, the concept of trust has been examined in various disciplines including human-robot interaction \cite{Sheridan:2016kn}, organizational management \cite{Mayer:1995ju}, psychology \cite{Klimoski1976}, and social philosophy \cite{Faulkner2017}. The diverse interest has generated multiple definitions of trust. In our study, we use one of the widely used definitions in human-autonomy interaction. Further research should consider alternative definitions from other disciplines. Please see \cite{Lewis2018,Mayer:1995ju} for more discussion on the definitions of trust. Second, the trust dynamics model used in the present study is primarily ability-centric. Even though the focus of the present study is not on the trust dynamics model, further research should be conducted to validate the simulation findings when a more inclusive trust dynamics model is available. Such dynamics models could consider additional characteristics of the trustee including the trustee's benevolence and integrity.}

\bibliographystyle{IEEEtran}
\bibliography{reference}


\end{document}